\documentclass[a4paper,twocolumn]{article}
\pdfoutput=1 
\usepackage{amssymb}
\usepackage{amsmath}
\usepackage{algorithm2e}
\usepackage{subcaption}
\usepackage{tabularx}
\usepackage[colorlinks,pdfusetitle]{hyperref}
\usepackage{graphicx}
\begin{document}

\title{Memory-efficient and fast implementation of local adaptive binarization methods \thanks{Declarations of interest: none}}
\author{Chungkwong Chan \thanks{School of Mathematics, Sun Yat-Sen University, 135 Xingang Xi Road, Guangzhou, 510275, China} \thanks{Email address: chsongg@mail2.sysu.edu.cn or chan@chungkwong.cc}}

\maketitle

\begin{abstract}
	Binarization is widely used as an image preprocessing step to separate object especially text from background before recognition. For noisy images with uneven illumination such as degraded documents, threshold values need to be computed pixel by pixel to obtain a good segmentation. Since local threshold values typically depend on moment-based statistics such as mean and variance of gray levels inside rectangular windows, integral images which are memory consuming are commonly used to accelerate the calculation. Observed that moment-based statistics as well as quantiles in a sliding window can be computed recursively, integral images can be avoided without neglecting speed, more binarization methods can be accelerated too. In particular, given a $H\times W$ input image, Sauvola's method and alike can run in $\Theta (HW)$ time independent of window size, while only around $6\min\{H,W\}$ bytes of auxiliary space is needed, which is significantly lower than the $16HW$ bytes occupied by the two integral images. Since the proposed technique enable various well-known local adaptive binarization methods to be applied in real-time use cases on devices with limited resources, it has the potential of wide application.
\end{abstract}

\begin{description}
	\item[Keywords] Adaptive binarization; Local thresholding; Space complexity; Time complexity; Optical character recognition
	\item[2010 MSC] 68U10; 68T45; 68W40
	
\end{description}

\section{Introduction\label{sec:intro}}

Binarization is an image processing procedure that convert a grayscale image into a binary image, it is commonly used to distinguish objects from background\cite{Mehmet}. In particular, binarization is widely adapted as a preprocessing step before optical character recognition(OCR), where textual components are first extracted from images. The quality of binarization is critical to the accuracy of OCR, especially for degraded documents\cite{GATOS2006317} and license plates\cite{4518951}.

A global binarization method divides pixels into foreground and background using a single threshold value, where the gray level of each pixel is compared to the threshold value. Since brightness and contrast are different from image to image, a good threshold value need to be computed per image. Most notably, Otsu\cite{4310076} suggested to choose a threshold value such that interclass variance of gray levels is maximized. 

A local binarization method assign possibly different threshold values to different pixels in an image. The threshold value for a pixel $(i,j)$ is calculated using statistics of the gray levels of pixels inside a neighborhood of $(i,j)$. In practice, the neighborhood is usually taken to be a $h\times w$ rectangular window centered at $(i,j)$, commonly used local statistics are mean $m_{ij}$ and standard deviation $s_{ij}$. For example, the threshold value for pixel $(i,j)$  of a $H\times W$ grayscale image $(I_{ij})_{i=0,\ldots,H-1; j=0,\ldots,W-1}$ is taken to be $$t_{ij}=m_{ij}(1+k(\frac{s_{ij}}{R}-1))$$ in the well known method proposed by Sauvola and Pietik\"{a}inen\cite{Sauvola2000Adaptive}, where $k$ is a tunable positive parameter and $R$ is the dynamic range of standard deviation. In most application, allowed gray levels are $\{0,\ldots,255\}$, so $255/2$ would be an upper bound of the standard deviation according to Popoviciu's inequality, thus $R$ is usually taken to be $128$. 

\begin{figure}
	\centering
	\begin{minipage}{\linewidth}
		\includegraphics[width=\linewidth]{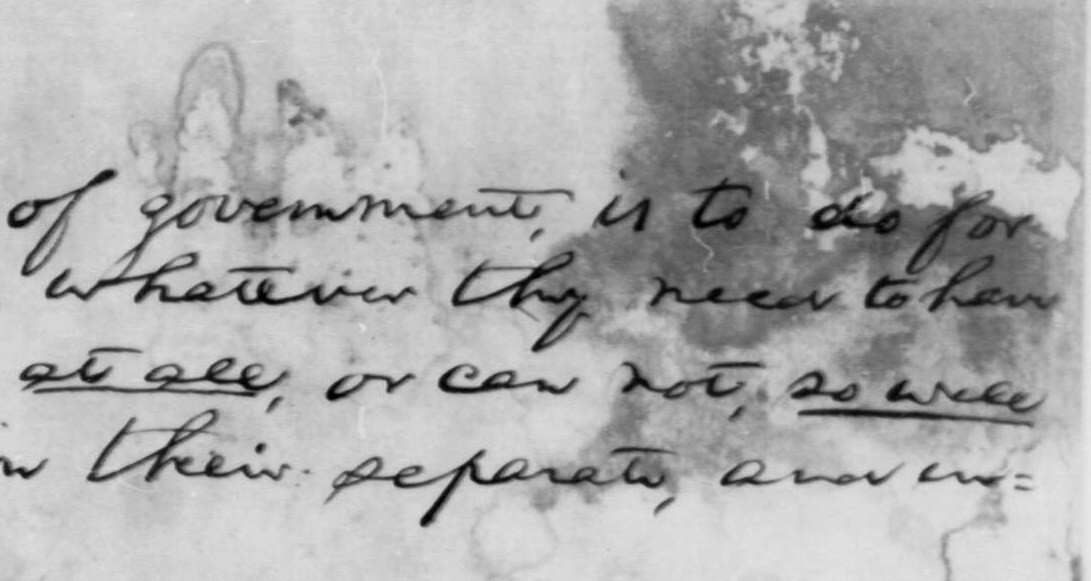}
		\subcaption{Before binarization}\label{fig:grayscale}
		\includegraphics[width=\linewidth]{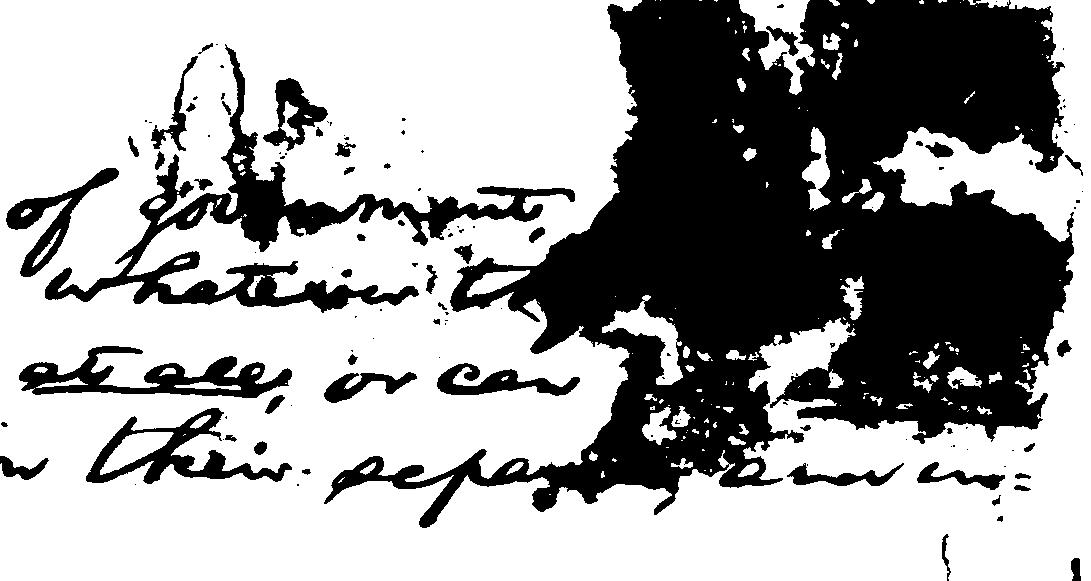}
		\subcaption{Binarized by Otsu's method}\label{fig:otsu}
		\includegraphics[width=\linewidth]{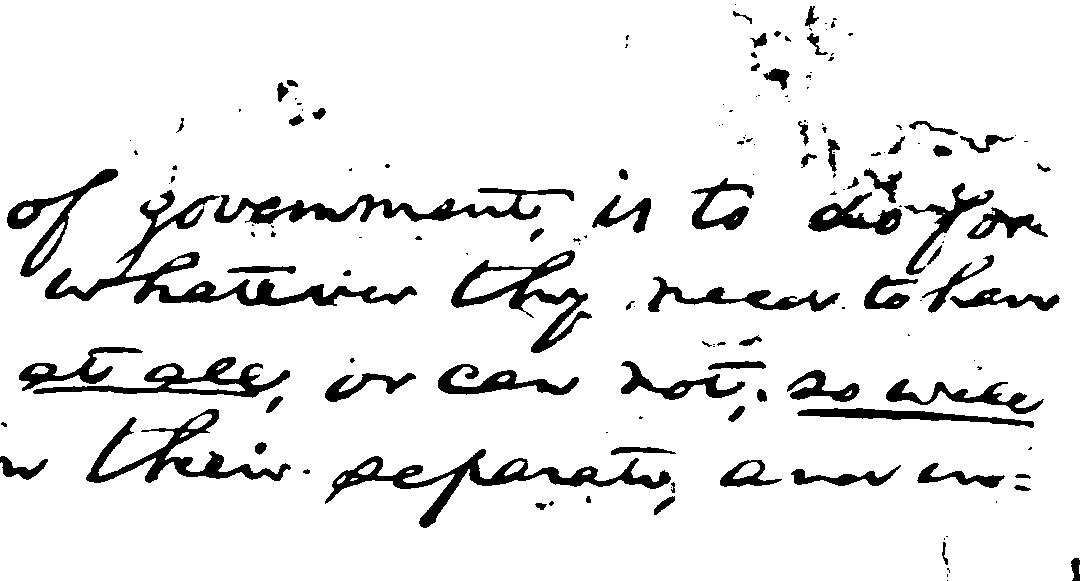}
		\subcaption{Binarized by Sauvola's method($k=0.5, h=w=32$)}\label{fig:sauvola}
	\end{minipage}
	\caption{Binarization of a sample image}\label{fig:binarization}
\end{figure}

Global binarization methods are fast, but they are not robust to noise and uneven illumination which are commonly seen. For example, a single threshold value may not be adequate to separate text from coffee stains. Fig. \ref{fig:binarization} compares the output of Otsu's method and that of Sauvola's method, where the input image is taken from the dataset of ICDAR 2009 Document Image Binarization Contest\cite{5277767}. In contrast, local binarization methods are computationally intensive. Since 
$$m_{ij}=\frac{1}{hw}\sum_{\substack{i-\frac{h}{2}<\alpha\leq i+\frac{h}{2}\\j-\frac{w}{2}<\beta\leq j+\frac{w}{2}}}I_{\alpha,\beta}$$ and $$s_{ij}=\sqrt{\frac{1}{hw}\sum_{\substack{i-\frac{h}{2}<\alpha\leq i+\frac{h}{2}\\j-\frac{w}{2}<\beta\leq j+\frac{w}{2}}}I_{\alpha,\beta}^2-m_{ij}^2}$$, the time complexity would be $\Theta (HWhw)$ if all the threshold values of Sauvola's method are computed directly using the definition, the speed decrease as the window size increase.

For real-time use cases such as mobile devices that help foreigners or visually impaired persons to understand encountered text\cite{1334351}, memory-efficient and fast implementations of local binarization methods are demanded. In this paper, such an implementation is presented, where a generally applicable recursive technique is proposed to compute quantities of the form $$\sum_{\substack{\alpha_0<\alpha\leq\alpha_1\\\beta_0<\beta\leq\beta_1}}f(I_{\alpha,\beta})$$. The technique can also be extends to compute various features of image which are useful for object detection and classification efficiently using limited memory.

The remainder of this paper is divided into four sections. Section \ref{sec:related} summarizes previous attempts to accelerate local adaptive binarization and their limitations. Section \ref{sec:proposed} describes the proposed implementation in details and analyzes it complexity. Section \ref{sec:expr} presents benchmark of the implementation. Section \ref{sec:conclusion} concludes the paper.

\section{Related works\label{sec:related}}

\begin{figure}
	\centering
	\includegraphics[width=\linewidth]{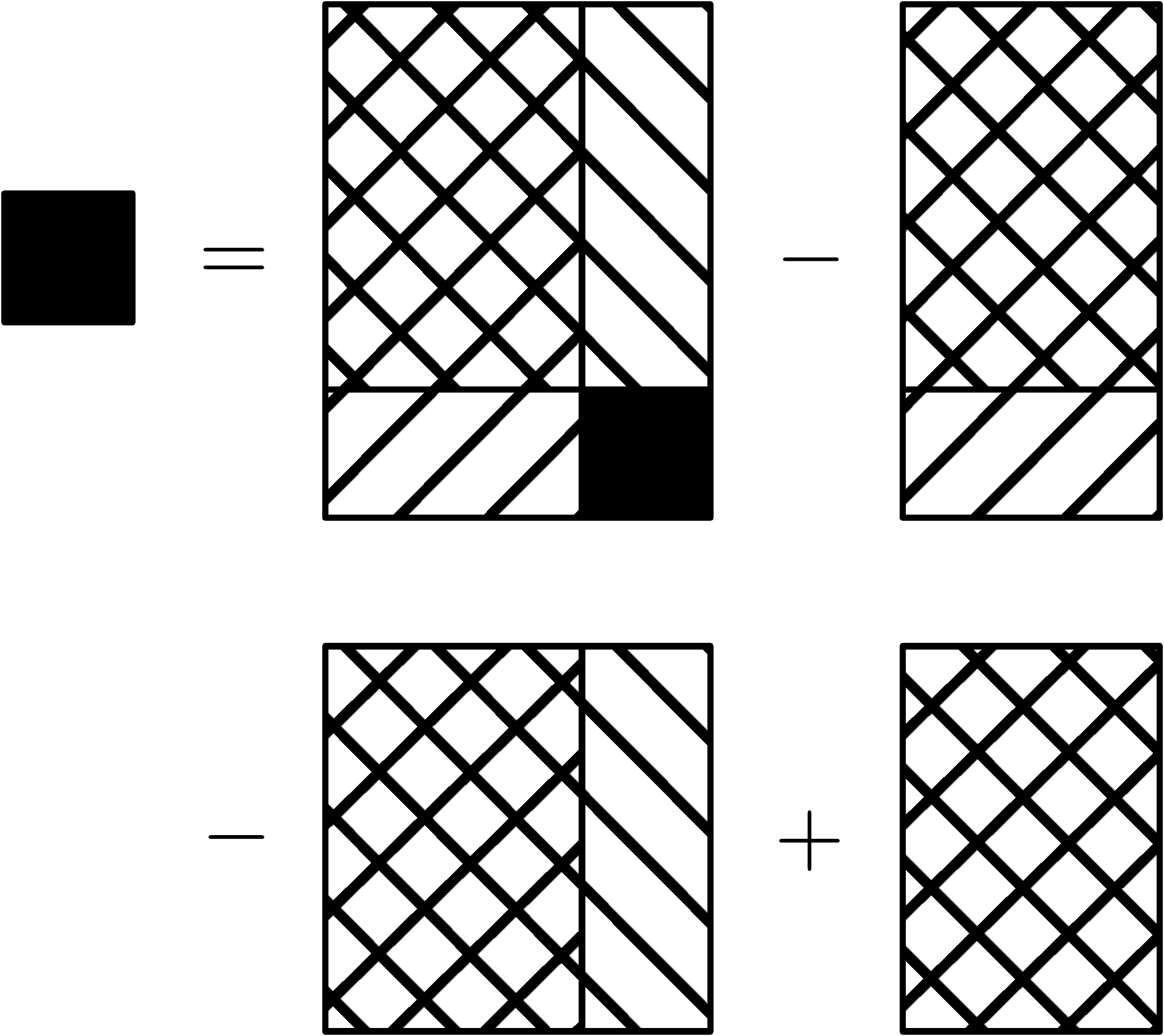}
	\caption{Calculation of sum inside a window given an integral image}\label{fig:integral}
\end{figure}

Bradley et al.\cite{Bradley} used integral images to speed up the calculation of the mean of gray levels inside a window. Shafait et al.\cite{Shafait} extended the trick to calculate the standard deviation. Let $J_{i,j}=\displaystyle\sum_{\substack{\alpha\leq i\\\beta\leq j}}f(I_{\alpha,\beta})$, then $\displaystyle\sum_{\substack{\alpha_0<\alpha\leq\alpha_1\\\beta_0<\beta\leq\beta_1}}f(I_{\alpha,\beta})=J_{\alpha_1,\beta_1}-J_{\alpha_0,\beta_1}-J_{\alpha_1,\beta_0}+J_{\alpha_0,\beta_0}$ as shown in Fig. \ref{fig:integral}. The integral image $(J_{ij})_{i=0,\ldots,H-1; j=0,\ldots,W-1}$ can be computed recursively using the following recurrence relation: $$J_{i,j}=\begin{cases}0 &, i<0 \textrm{ or } j<0\\J_{i-1,j}+J_{i,j-1}+f(I_{i,j})\\ -J_{i-1,j-1} & ,\textrm{otherwise}\end{cases}$$. Therefore, the time complexity of Sauvola's method can be reduced to $\Theta (HW)$, which is independent of window size. However, consumption of memory increased significantly from $\Theta (1)$ to $\Theta (HW)$ because the two additional integral images need to be stored.

Hardware acceleration is also considered. Chen et al.\cite{Chen2015} proposed a parallel implementation of Sauvola's method using GPU. Najafi et al.\cite{7104144} presented a full hardware implementation using stochastic circuits.

\section{Proposed implementation\label{sec:proposed}}

In this section, a memory-efficient and fast implementation of Sauvola's method is proposed, no specialized hardware is required. The novel technique can also be applied to speed up a wide range of local adaptive binarization methods while memory requirement is moderate.

\subsection{Overview}

\begin{figure}
	\centering
	\begin{minipage}{\linewidth}
		\includegraphics[width=\linewidth]{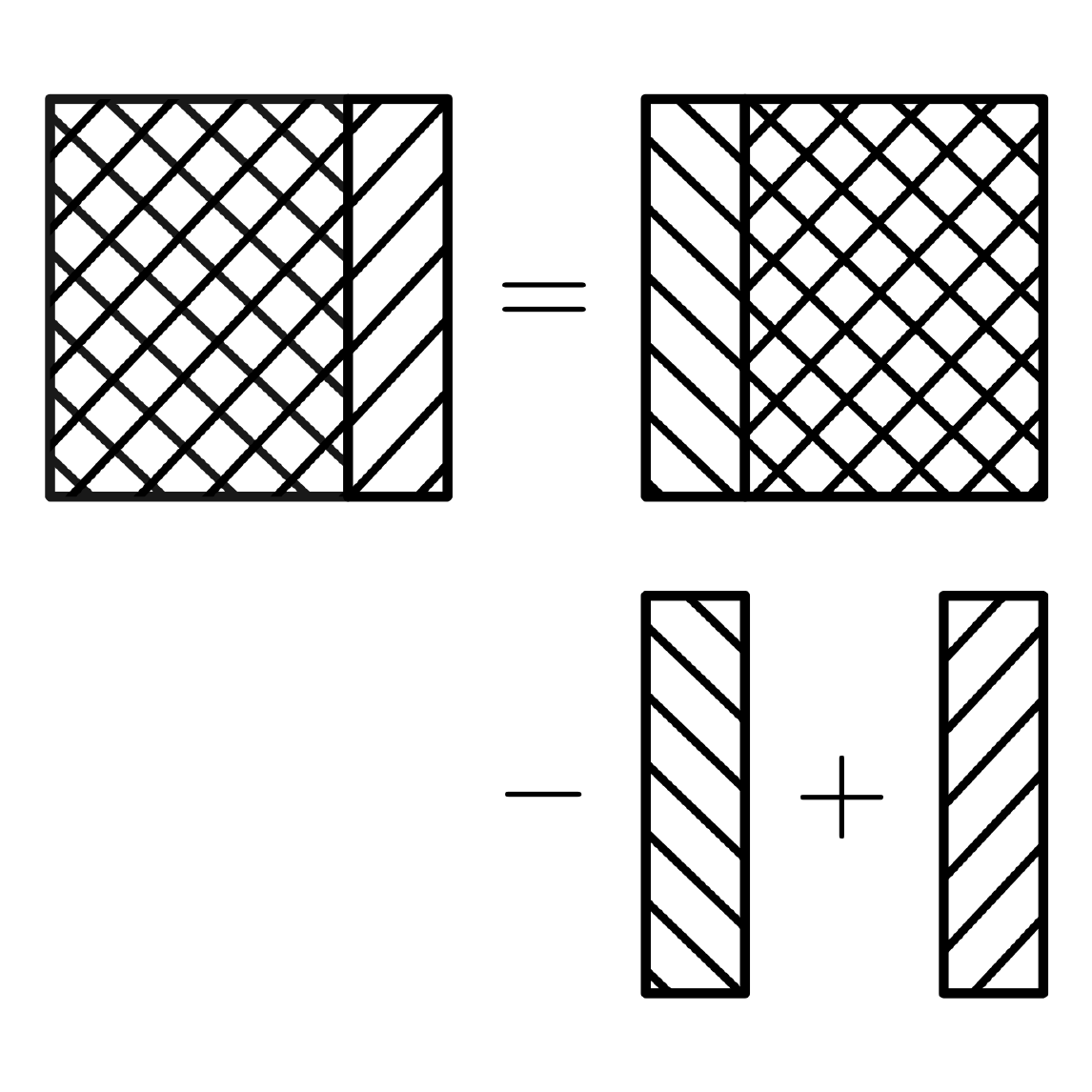}
		\subcaption{Window level}\label{fig:window}
		\includegraphics[width=\linewidth]{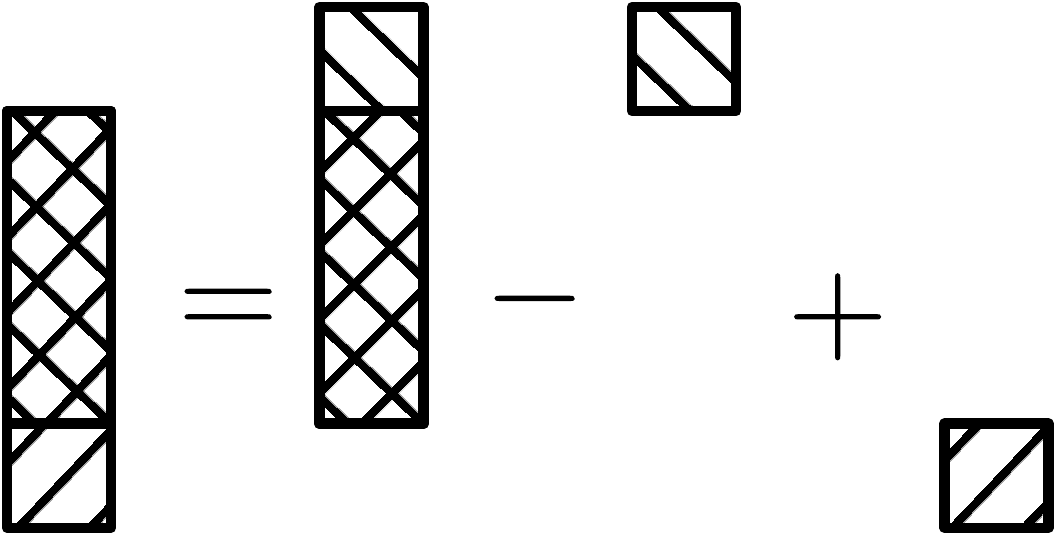}
		\subcaption{Stripe level}\label{fig:stripe}
	\end{minipage}
	\caption{Calculation of sum inside a window given that of a window next to it}\label{fig:diff}
\end{figure}

The primary observation is that the sum of $f$ of gray levels inside a window can be computed using the one next to it as shown in Fig. \ref{fig:diff}, more formally,
\begin{align*}\sum_{\substack{\alpha_0<\alpha\leq\alpha_1\\\beta_0<\beta\leq\beta_1}}f(I_{\alpha,\beta})=&\sum_{\substack{\alpha_0<\alpha\leq\alpha_1\\\beta_0-1<\beta\leq\beta_1-1}}f(I_{\alpha,\beta})\\ +&\sum_{\alpha_0<\alpha\leq\alpha_1}f(I_{\alpha,\beta_1})\\ -&\sum_{\alpha_0<\alpha\leq\alpha_1}f(I_{\alpha,\beta_0})\end{align*} for integers $\alpha_0, \alpha_1, \beta_0\text{ and }\beta_1$, where \begin{align*}\sum_{\alpha_0<\alpha\leq\alpha_1}f(I_{\alpha,\beta})=\sum_{\alpha_0-1<\alpha\leq\alpha_1-1}f(I_{\alpha,\beta})&+f(I_{\alpha_1,\beta})\\ &-f(I_{\alpha_0,\beta})\end{align*}. Therefore, if the threshold values of pixels are computed in row-major order, the implementation only need to keep track of the quantities $\displaystyle\sum_{i-\lfloor\frac{h+1}{2}\rfloor<\alpha\leq i+\lfloor\frac{h}{2}\rfloor}f(I_{\alpha,\beta})$ where $\beta=0, \ldots, W-1$.

\subsection{Algorithm}

Given the observation, Sauvola's method can be implemented as in Algorithm \ref{algo:sauvola}. For simplicity, the following assumptions are made:

\begin{itemize}
	\item All undefined elements(due to index out of bounds) of arrays have a value zero;
	\item The input image and the output image do not overlap in memory, otherwise $I_{i-o,j}$ may be overwritten unintentionally. The algorithm can be modified to allow in-place binarization, but additional space is needed to store up to $oW$ gray levels.
\end{itemize}

\begin{algorithm}
	\SetKwData{Left}{left}\SetKwData{This}{this}\SetKwData{Up}{up}
	\SetKwFunction{Union}{Union}\SetKwFunction{FindCompress}{FindCompress}
	\SetKwInOut{Input}{input}\SetKwInOut{Output}{output}
	\Input{A grayscale bitmap image $\mathbf{I}$ with width $W$ and height $H$, the width $w$ and height $h$ of the window, a nonnegative parameter $k$, the range of standard deviation $R$}
	\Output{A binary bitmap image $\mathbf{I}'$}
	\BlankLine
	$l\leftarrow \lfloor \frac{w+1}{2} \rfloor$\;
	$r\leftarrow \lfloor \frac{w}{2} \rfloor$\;
	$o\leftarrow \lfloor \frac{h+1}{2} \rfloor$\;
	$u\leftarrow \lfloor \frac{h}{2} \rfloor$\;
	\For{$j\leftarrow 0$ \KwTo $W-1$}{
		$C_j\leftarrow 0$\;
		$D_j\leftarrow 0$\;
		\For{$i\leftarrow 0$ \KwTo $u-1$}{
			$C_j\leftarrow C_j+I_{i,j}$\;
			$D_j\leftarrow D_j+I_{i,j}^2$\;
		}
	}
	\For{$i\leftarrow 0$ \KwTo $H-1$}{
		\For{$j\leftarrow 0$ \KwTo $W-1$}{
			$C_j\leftarrow C_j+I_{i+u,j}-I_{i-o,j}$\;
			$D_j\leftarrow D_j+I_{i+u,j}^2-I_{i-o,j}^2$\;
		}
		$c\leftarrow 0$\;
		$d\leftarrow 0$\;
		\For{$j\leftarrow 0$ \KwTo $r-1$}{
			$c\leftarrow c+C_j$\;
			$d\leftarrow d+D_j$\;
		}
		\For{$j\leftarrow 0$ \KwTo $W-1$}{
			$n\leftarrow (\min\{j+r,W-1\}-\max\{j-l,-1\})\times (\min\{i+u,H-1\}-\max\{i-o,-1\})$\;
			$c\leftarrow c+C_{j+r}-C_{j-l}$\;
			$d\leftarrow d+D_{j+r}-D_{j-l}$\;
			$m\leftarrow\frac{c}{n}$\;
			$v\leftarrow\frac{d}{n}-m^2$\;
			\tcc{Change the following condition when you are implementing another binarization method}
			\eIf{$I_{ij}+m(k-1)\leq 0\vee(I_{ij}+m(k-1))^2\leq \frac{k^2m^2v}{R^2}$}{$I'_{ij}\leftarrow 0$}{$I'_{ij}\leftarrow 1$}
		}
	}
	\caption{Sauvola's method}\label{algo:sauvola}
\end{algorithm}

It should also be noted that $I_{ij}\leq m_{ij}(1+k(\frac{s_{ij}}{R}-1))$ is equivalent to $I_{ij}+m_{ij}(k-1)\leq 0$ or $(I_{ij}+m_{ij}(k-1))^2\leq \frac{k^2m_{ij}^2s_{ij}^2}{R^2}$ since $k\geq 0$, so computation of square root can be avoided to save time. 

\subsection{Complexity}

It is straightforward to see from Algorithm \ref{algo:sauvola} that $\Theta (W)$ auxiliary space is needed, where the space occupied by the input image and the output image is not taken into account. By exchanging the two axis, one may implement a version demanding $\Theta (H)$ auxiliary space. Therefore, the requirement of auxiliary space can be further decreased to $\Theta (\min\{H,W\})$ by choosing one of the two versions per image, it is much smaller than $\Theta (HW)$ which is needed for integral images. In fact, $\min\{H,W\}\leq\sqrt{HW}\ll HW$. 

In addition, shorter integer types can be used to store intermediate quantities since they are much smaller than the elements of integral images. In the usual scenario, gray levels range from 0 to 255 and the side of windows rarely exceed 257, $255\times 257=65535=2^{16}-1$ and $255\times 255\times 257<2^{32}-1$ by direct calculations, it follows that an element of $C$ only need to occupy an unsigned 16-bit integer and an element of $D$ only need to occupy a 32-bit integer. In contrast, a page of A4 document scanned at 600DPI contains up to $8.27\times 11.7 \times 600^2=34833240>2^{32}/255$ pixels, so an element of an integral image need to occupy a 64-bit integer to prevent overflow.

It is straightforward to see from Algorithm \ref{algo:sauvola} that its time complexity is $\Theta (HW)$ and does not depend on window size, which is the same as the algorithm based on integral images. Since reduced memory usage leads to a higher cache hit rate, the speed may increase slightly.

\subsection{Generalization}

Most of the local thresholding methods derived from Niblack's method only rely on mean and variance of gray levels inside rectangular windows, Table \ref{tab:direct} listed a few of them, consult a survey\cite{Saxena2019} for even more examples. Therefore, Algorithm \ref{algo:sauvola} can be adapted to any of them by simply modifying the condition in the ``if'' statement. Obviously, the analysis on time complexity and space complexity remains unchanged.

\begin{table}
	\caption{Binarization methods that can be implemented by changing a single condition in Algorithm \ref{algo:sauvola}}
	\label{tab:direct}
	\begin{tabularx}{\linewidth}{p{0.3\linewidth}X}
		\hline\noalign{\smallskip}
		Method & Threshold value \\
		\noalign{\smallskip}\hline\noalign{\smallskip}
		Niblack\cite{Niblack} & $m_{ij}+ks_{ij}$ \\
		Sauvola et al.\cite{Sauvola2000Adaptive} & $m_{ij}(1+k(\frac{s_{ij}}{R}-1))$ \\
		Wolf et al.\cite{1048482} & $m_{ij}-k(m_{ij}-L)(1-\frac{s_{ij}}{R})$ \\
		Feng et al.\cite{Feng} & $\alpha_1m_{ij}+k_1(\frac{s_{ij}}{R})^{1+\gamma}(m_{ij}-L)+k_2(\frac{s_{ij}}{R})^{\gamma}L$\\
		Rais et al.\cite{1492847}&$m_{ij}+0.3\frac{m_{ij}s_{ij}-MS}{\max\{m_{ij}s_{ij},MS\}}s_{ij}$\\
		Khurshid et al.\cite{Khurram}&$m_{ij}+k\sqrt{s_{ij}^2+m_{ij}^2\frac{hw-1}{hw}}$\\
		Phansalkar et al.\cite{5739305}&$m_{ij}(1+pe^{-qm_{ij}}+k(\frac{s_{ij}}{R}-1))$\\
		\noalign{\smallskip}\hline
	\end{tabularx}
	\scriptsize
	Notations:
	\begin{description}
		\item[$L$] global minimum of gray levels
		\item[$M$] global mean of gray levels
		\item[$S$] global standard deviation of gray levels
	\end{description}
	\normalsize
\end{table}

Some of the local thresholding methods used local maximum and/or minimum. For example, Bernsen's method\cite{Bernsen86dynamicthresholding} used midrange as threshold value. It is possible to calculate local maximum and minimum recursively using deques. Although $\Theta (HW)$ time and $\Theta (\min\{Wh,Hw\})$ auxiliary space would be needed in the present situation, it should remain much faster than the definitive implementation, while consumption of memory is acceptable since windows are small compared with input images. The same technique can also speed up binarization methods that estimate local contrast by range of gray levels\cite{6373726}.

In general, local quantiles such as median of gray levels can also be computed recursively by maintaining histograms. $\Theta (HW\max\{h,w\})$ time and $\Theta (\min\{W,H\})$ auxiliary space would be needed in this case, depending on the number of gray levels allowed.

Compound binarization methods which are based on a supported adaptive thresholding method can benefit from the proposed technique too. Gatos et al.\cite{GATOS2006317} added preprocessing and postprocessing steps before and after Sauvola's thresholding to reduce noise. Chamchong et al.\cite{5212718} modified threshold values which are outliers. Moghaddam et al.\cite{FARRAHIMOGHADDAM20102186} and Lazzara et al.\cite{Lazzara2014} applied underlying local thresholding multiple times to downsampled images.

The proposed technique may be of interest beyond binarization since it can replace the memory-hogging integral images without neglecting speed in many cases. As a potential application, Haar-like features\cite{Viola2004} which are widely adapted in object detection can be computed recursively, so one may implement real-time face detection on devices with limited memory. Here is an outline: a small number of features are computed recursively to produce candidates of face first, then remaining features are computed using local integral images of those regions to classify them.

\section{Experiment\label{sec:expr}}

\begin{figure}
	\centering
	\includegraphics[width=\linewidth]{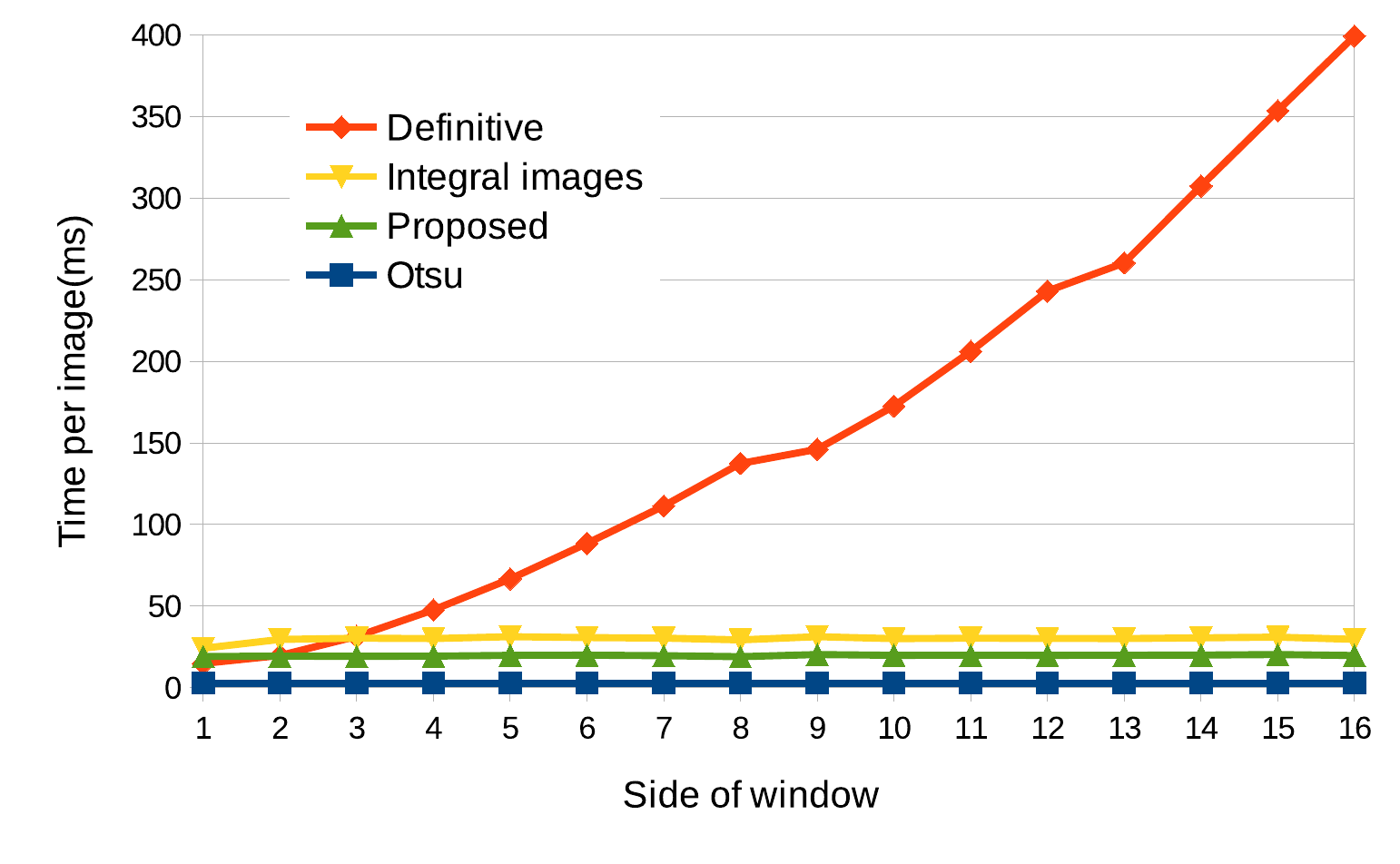}
	\caption{Comparison of CPU time usage}\label{fig:time}
\end{figure}

In order to check the effectiveness of the proposed technique, the three implementations\footnote{The source code can be found at \url{https://github.com/chungkwong/binarizer}.} of Sauvola's method are tested on the dataset from ICDAR 2013 Document Image Binarization Contest\cite{6628857} together with Otsu's method. The three implementations of Sauvola's method produced identical binarized images for each input image and window size tested, so the correctness of the implementations is confirmed. Fig. \ref{fig:time} verified the speed of the proposed implementation is independent of window size. The proposed implementation is about 30\% faster than the approach using integral images, but still approximately six times slower than Otsu's method.

\section{Conclusion\label{sec:conclusion}}

A memory-efficient and fast implementation of Sauvola's method is presented. A $H\times W$ grayscale image can be binarized in $\Theta (HW)$ time independent of window size, while the definitive implementation needs $\Theta (HWhw)$ time for $h\times w$ window. Meanwhile, only $\Theta (\min\{H,W\})$ space is occupied by intermediate data structures, while the approach using integral images requires $\Theta (HW)$. The proposed technique can also benefit most of the local adaptive thresholding methods, whenever local statistics like mean, standard deviation or quantiles are concerned. Therefore, a large class of widely used local adaptive binarization methods are made available to real-time use cases on devices with limited resources.

\bibliographystyle{main} 
\bibliography{main}

\end{document}